\documentclass[a4paper,fleqn]{cas-dc}

\usepackage[authoryear,longnamesfirst]{natbib}
\usepackage{balance}
\let\printorcid\relax 
\usepackage{blindtext}
\usepackage{stfloats}
\def\tsc#1{\csdef{#1}{\textsc{\lowercase{#1}}\xspace}}
\tsc{WGM}
\tsc{QE}

\begin{document}
\let\WriteBookmarks\relax
\def\floatpagepagefraction{1}
\def\textpagefraction{.001}

\shorttitle{Synthetic Data Generation for Improved Pain Recognition in Videos}    

\shortauthors{Nasimzada et al.}  

\title [mode = title]{Towards Synthetic Data Generation for Improved Pain Recognition in Videos under Patient Constraints}  



%

\author[1]{Jonas Nasimzada}
\author[2]{Jens Kleesiek}
\author[1]{Ken Herrmann}
\author[1]{Alina Roitberg$^\dagger$}
\author[2]{Constantin Seibold$^\dagger$}

\cormark[2]

\fnmark[1]



\affiliation[1]{organization={University Stuttgart, Institute for AI},
            addressline={Universitätsstraße 32 }, 
            city={Stuttgart},
            postcode={70569}, 
            state={Baden-Württemberg},
            country={Germany}}


\affiliation[2]{organization={University Clinic Essen},
            addressline={Giradetstraße 2}, 
            city={Essen},
            postcode={45131}, 
            state={North-Rhine-Westphalia},
            country={Germany}}

\cortext[1]{corresponding author. (E-Mail: \texttt{constantin.seibold@uk-essen.de})\\
All author emails: \texttt{st171793@stud.uni-stuttgart.de} (J. Nasimzada); \texttt{jens.kleesiek@uk-essen.de} (J. Kleesiek); \texttt{ken.herrmann@uk-essen.de} (K. Herrmann); \texttt{alina.roitberg@ki.uni-stuttgart.de} (A. Roitberg); \texttt{constantin.seibold@uk-essen.de} (C. Seibold)
}


\nonumnote{$^\dagger$ equal senior co-authorship.}


\begin{abstract}
Recognizing pain in video is crucial for improving patient-computer interaction systems, yet traditional data collection in this domain raises significant ethical and logistical challenges. 
This study introduces a novel approach that leverages synthetic data to enhance video-based pain recognition models, providing an ethical and scalable alternative. 
We present a  pipeline that synthesizes realistic 3D facial models by capturing nuanced facial movements from a small participant pool, and mapping these onto diverse synthetic avatars. This process generates 8,600 synthetic faces, accurately reflecting genuine pain expressions from varied angles and perspectives.

Utilizing advanced facial capture techniques, and leveraging public datasets like CelebV-HQ and FFHQ-UV for demographic diversity, our new synthetic dataset significantly enhances model training while ensuring privacy by anonymizing identities through facial replacements. 

Experimental results demonstrate that models trained on combinations of synthetic data paired with a small amount of real participants achieve superior performance in pain recognition, effectively bridging the gap between synthetic simulations and real-world applications. Our approach addresses data scarcity and ethical concerns, offering a new solution for pain detection and opening new avenues for research in privacy-preserving dataset generation. All resources are publicly available to encourage further innovation in this field.
\end{abstract}





\maketitle

\section{Introduction}\label{sec:introduction}
Pain significantly impairs the daily life of millions of people  with up to 11.2 percent of American population experiencing significant pain on a daily basis~\citep{nahin2015estimates}. Similarly, in 2021, 20.9\% of U.S. adults experienced chronic pain, with 6.9\% enduring high-impact chronic pain that significantly restricts daily activities~\citep{rikard2023chronic}, whereas. If untreated, such pain conditions, both acute and chronic, can lead to massive loss in economy productivity~~\citep{phillips2009cost}. To minimize pain across the population, it is of utmost importance to identify scenarios in which pain can occur not only in patient care~\citep{matthias2010patient}, but also in daily living~\citep{resnick2019pain} and working scenarios~\citep{cote2008burden}.

While automated video analysis shows potential in various aspects of understanding human actions~\citep{kay2017kinetics,
carreira2017quo} and emotions~\citep{soleymani2011multimodal,ebrahimi2015recurrent}, pain recognition poses a rather difficult issue due to the difficulty of dataset aggregation. Regulations such as the GDPR~\citep{gdpr2016} make it difficult to gather pain in the wild. Capturing spontaneous pain episodes in real-world situations is difficult and costly, requiring researchers to follow participants in their natural environments or use wearable recording devices~\citep{Patel_Doku_Tennakoon_2003}. As this is a major intrusion in the participants personal life, such a study must pass a rigorous ethics reviews~\citep{price2019privacy}. Similarly, if one were to construct such a dataset under a controlled setting, participants have to undergo considerable deliberate pain. Designing such a study in an ethical manner extremely and finding willing participants in a non-exploitable social situation can pose a challenge. As such, publicly available datasets as well as the number of subjects on this topic are limited~\citep{zhang2013high, walter2013biovid}. In return, this data scarcity might lead data-hungry networks, which are prevalent in video processing~\citep{monfort2019moments,carreira2017quo},  not only be to be lacking in performance but also be fragile in face of new scenarios such as underrepresented races, sex, or locations~\citep{schiappa2023large,dooley2021robustness,karkkainen2021fairface}.

\textit{How can we enable dataset sizes that lead to increased performance and robustness of neural networks without running into the same ethical consideration?}

\begin{figure*}
    \centering
    \includegraphics[width=\linewidth]{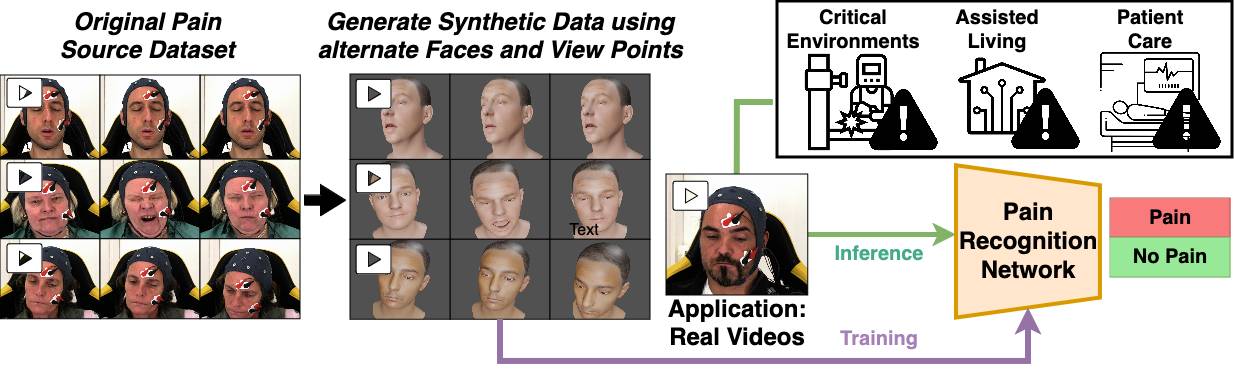}
    \caption{We leverage facial features extracted from a small dataset with real recordings with people in pain for generating a synthetic dataset with diverse faces and viewpoints to address logistical, ethical and bias-related challenges of such data collection.}
    \label{fig:graphical_abstract}
\end{figure*}

\noindent A prevalent path to generate additional in many areas of computer vision is the simulation of the world to generate a digital twin for various scenarios like driving~\citep{ros2016synthia, richter2016playing} or daily living~\citep{roitberg2021let}. In these settings, the task that has to be solved is well understood by the vast majority of the population. When diving into more specialized scenarios, rough simulations such as games become difficult to utilize as the problems themselves are topics of active research, as is the case with the manifestation of pain~\citep{cowen2015assessing}. While facial expressions that are supposed to represent pain can be identified a variety of video games, the origin of the expression is derived from actors rather than a real stimulus. Hence, to synthetically create real facial movements in painful situations, rather than to rely on acted pain expressions, we propose the use of a pipeline which 
utilizes a small set of participants, captures their facial 3D movements and translates these onto a diverse set of faces. This allows us to enhance gender and ethnic diversity while also introducing perspectives from previously unseen viewpoints. In addition to these benefits, such a pipeline also enables further perspectives in regards to privacy preserving dataset generation, as the identities of the participants can be fully hidden by replacing their faces with people of the public domain. 

This study created an efficient pipeline for human mesh and texture generation, resulting in a dataset of 8,600 synthetic heads generated per perspective and texture. The datasets encompass varying facial textures and perspectives. We utilize different combinations of dataset subset, facial textures, and perspectives for the training of video networks to assess the feasibility of synthetic data for pain recognition in real data. We show that the use of synthetic data gained from this pipeline can significantly improve the performance of video-based pain recognition models. 

We estimate that our pipeline offers not only an alternative to pain recognition based on data
collection under immense ethical considerations but also provides potential perspectives on privacy-preserving dataset collection under facial requirements.  We hope creates a new avenue for future research focusing on capturing the essence of human behaviour in both, synthetic and real domains. All our code is publicly available at \href{https://github.com/JonasNasimzada/LetsPlay4Emotion}{github}.

\section{Related Work}\label{sec:related_work}
\subsection{Pain Recognition Methods}
Early approaches to pain classification often relied on precomputed features extracted from facial images or videos. Martinez et al. ~\citep{https://doi.org/10.48550/arxiv.1706.07154} utilized facial landmarks and Recurrent Neural Networks (RNNs) to estimate Prkachin and Solomon pain intensity (PSPI) levels. They employed personalized Hidden Conditional Random Fields (HCRFs) for individual Visual Analog Scale (VAS) assessments.

With the rise of deep learning, Convolutional Neural Networks (CNNs) have become prevalent for end-to-end pain recognition. Haque et al. ~\citep{8373837} proposed a method using RGB, depth, and thermal images to classify pain levels, leveraging a 2D CNN for feature extraction and Long Short-Term Memory (LSTM) networks for temporal analysis. Rodriguez et al. ~\citep{7849133} employed CNNs to extract features from the VGG-Faces dataset, followed by an LSTM for binary pain classification, utilizing the UNBC–McMaster pain database. Wang et al. ~\citep{8296449} adopted a face verification network trained on the WebFace dataset, fine-tuned for pain estimation as a regression problem.

While these methods demonstrate the effectiveness of CNN-LSTM architectures, they primarily depend on pre-existing datasets, which may lack diversity and introduce biases. Our work diverges by implementing 3D CNNs for video-based pain recognition, leveraging synthetic data to supplement existing datasets, thus enhancing model robustness and generalizability.

\subsection{Challenges in Pain Dataset Collection}
\begin{table*}[b]
    \centering
    \caption{Adult Pain Recognition Databases. $\star$ denotes no longer available datasets. $\dagger$ denotes datasets that are not specifically focused on pain. UNBC-McMaster SPEA~\citep{lucey2011unbc}
BioVid Heat ~\citep{walter2013biovid}
BP4D-Spontaneous~\citep{zhang2014bp4d}
BP4D+~\citep{zhang2016bp4dplus}
MIntPAIN~\citep{tobon2020mintpain}}
    \begin{tabular}{lccccc}
        \toprule
        \textbf{Dataset} & \textbf{Age} & \textbf{Status} & \textbf{Stimulus}&\textbf{Participants} & \textbf{Origin}  \\
        \midrule
        UNBC-McMaster SPEA$\star$ &  - & shoulder pain & - & 25 & - \\
        BioVid Heat  &   20-65 years & healthy & Heat & 90 & European \\
        BP4D-Spontaneous$\dagger$ &   18-29 years & healthy & Cold & 41 & Diverse \\
        BP4D+$\dagger$ &    18-66 years & healthy & Cold & 140 & Diverse \\
        MIntPAIN &   22-42 years & healthy & Electrity& 20 & - \\
        \bottomrule
    \end{tabular}
    \label{tab:datasets}
\end{table*}

Collecting pain datasets presents significant ethical and logistical challenges. Conducting such studies often requires invasive methods, including following participants or using wearable recording devices, necessitating rigorous ethical review processes ~\citep{Patel_Doku_Tennakoon_2003}. Controlled experiments, on the other hand, involve administering deliberate pain, raising ethical concerns and participant recruitment challenges.

These factors contribute to the scarcity of publicly available pain datasets, limiting the diversity and volume of data necessary for training deep learning models ~\citep{zhang2013high, walter2013biovid, Zhang_2016_CVPR}. Consequently, data scarcity can hinder model performance, making them fragile in scenarios involving underrepresented demographics. We display currently existing datasets for pain recognition in Table\ref{tab:datasets}. Here we can notice the general limitation in number of participants. 

Our work addresses these challenges by employing synthetic data generation techniques, creating datasets that enhance demographic diversity and provide robust training sources for pain recognition models.

\subsection{Synthetic data for Pain Recognition}
Synthetic data in computer vision has gained traction, providing scalable solutions for data augmentation and training. Previous studies have successfully utilized synthetic environments for various tasks. Richter et al. ~\citep{https://doi.org/10.48550/arxiv.1608.02192,https://doi.org/10.48550/arxiv.1709.07322} exploited the Grand Theft Auto V (GTA V) environment to gather data for object detection, segmentation, and optical flow. Similarly, Krähenbühl ~\citep{8578410} used DirectX 11 to extract ground truth from video games for tasks like depth estimation and semantic segmentation.

\cite{https://doi.org/10.48550/arxiv.2107.05617} leveraged the life simulation game THE SIMS 4 to generate training examples for Activities of Daily Living (ADL) recognition. These studies underscore synthetic data's potential in scenarios with limited real-world data availability. However, applying synthetic data to pain recognition remains underexplored.

Despite its potential, synthetic data for pain recognition is still emerging. Prior work by Pikulkaew et al. ~\citep{9415827} used Generative Adversarial Networks (GANs) to create synthetic facial expressions but faced challenges in achieving realistic representations. These limitations underscore the complexity of accurately modeling pain expressions and the need for innovative approaches to synthetic data generation.

Our work addresses these challenges by capturing detailed 3D facial movements and translating them into diverse facial textures, creating a synthetic dataset that mirrors pain expressions across demographics. This method facilitates the training of robust pain recognition models and offers insights into privacy-preserving dataset generation. By synthesizing human behavior in synthetic domains, our contributions pave the way for future research in ethical and scalable pain recognition under  privacy considerations.

\section{Synthetic Dataset Generation}\label{sec:pipeline}

\begin{figure*}
    \centering
    \includegraphics[width=0.9\linewidth]{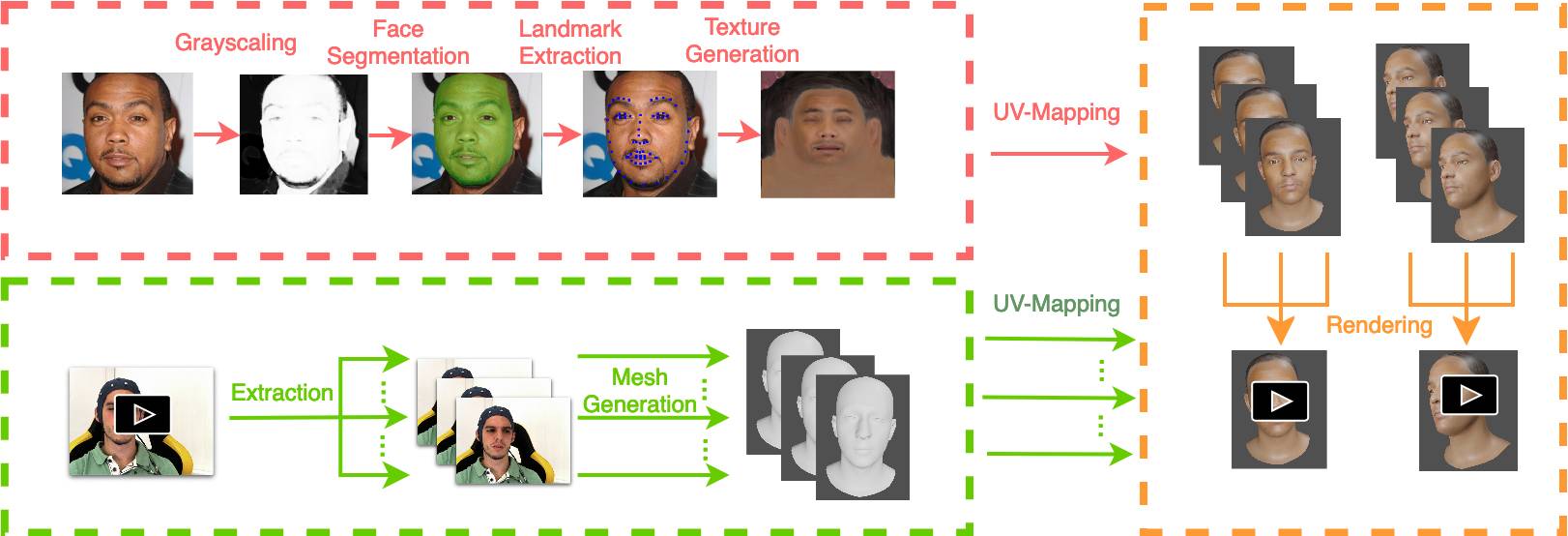}
    \caption{Generation Pipeline of Synthetic Head Videos. We extract facial textures from public domain faces for later translation. In parallel, we extract 3D volumes for each frame of the pain videos. Finally, we map the facial textures onto each frame-wise volume and render a scene from different perspectives. }
    \label{fig:pipeline}
\end{figure*}

Our pipeline for generating synthetic head videos builds on existing datasets such as the BioVid Heat Pain Database \cite{6617456}. Our approach integrates advanced methods in 3D mesh generation, texture mapping, and rendering to produce a diverse set of synthetic head videos for pain recognition research. The outline of this process is displayed in \ref{fig:pipeline}.

\subsection{Mesh Extraction}
Our pipeline begins with the generation of 3D head meshes from input video frames using the EMOCA framework \cite{Bai_2023_CVPR}. For each video frame  the head region is extracted via segmentation. The EMOCA framework processes these frames to produce consistent 3D head meshes. This process involves encoding each frame into context vectors that capture critical information about facial expressions, geometry, and pose. These context vectors are then used by a decoder to reconstruct detailed 3D head meshes. The resulting meshes are highly accurate and serve as the foundation for our synthetic head videos.

\begin{figure}[b]
    \centering
    \includegraphics[width=\linewidth, trim=0 450 0 0, clip]{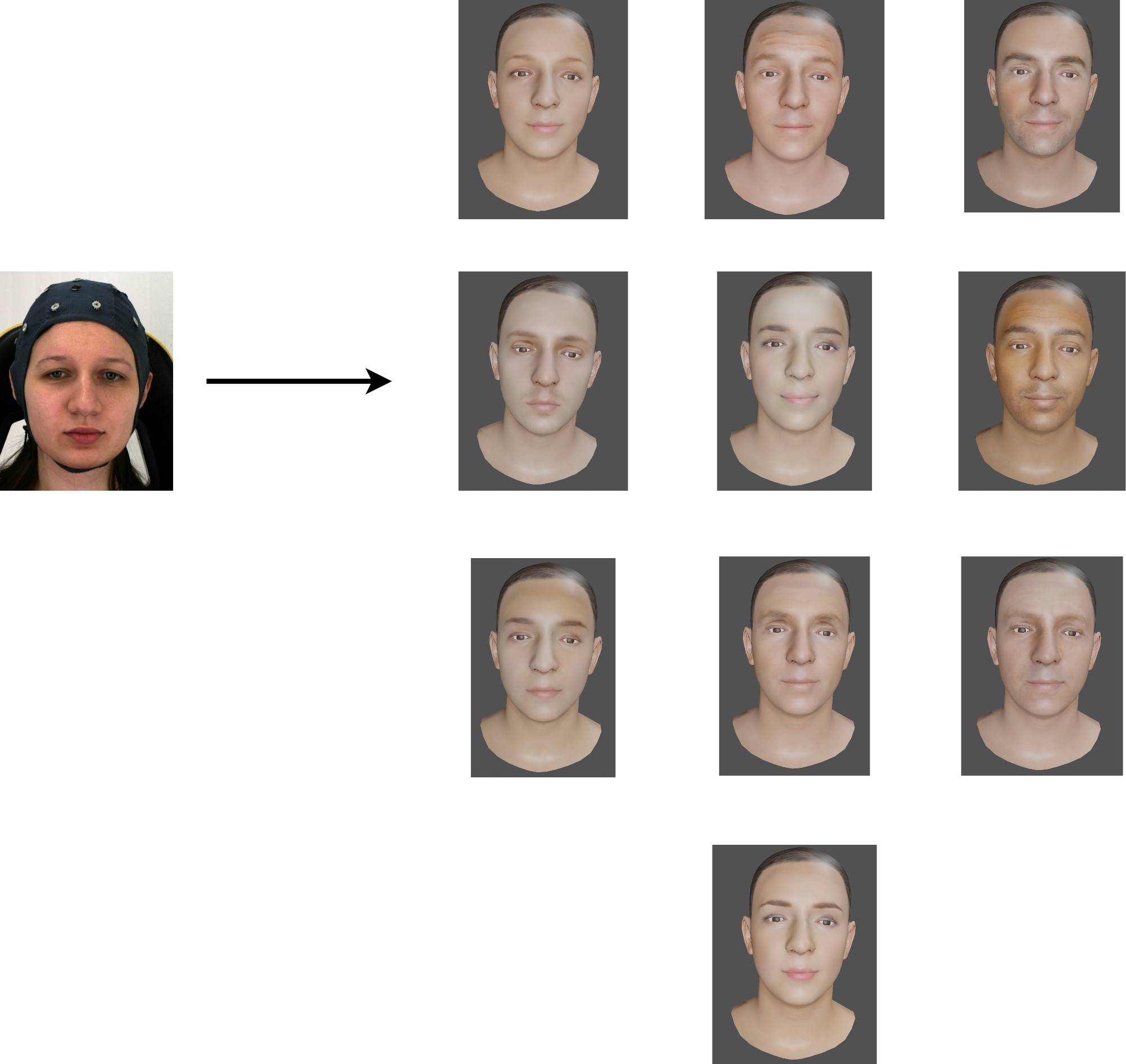}
    \caption{Example of different textures that are translated onto a participant of the BioVidHeat dataset.}
    \label{fig:texture_samples}
\end{figure}

\subsection{Texture Synthesis}
Texture mapping is performed via the FFHQ-UV repository \cite{Bai_2023_CVPR}, which provides a robust framework for generating high-resolution facial textures. The process begins with the detection of facial landmarks to ensure accurate texture application. A synthetic face is initially created from these landmarks, and high-resolution textures are then applied to this base mesh. To address common issues such as blank eye textures, we incorporate additional mapping refinements that enhance the realism of the synthetic heads. These refinements are essential for creating diverse and visually compelling head textures that reflect variations in age, gender, and ethnicity, as sourced from CelebV-HQ \cite{zhu2022celebvhq}. We display how final translations might look in Fig. \ref{fig:texture_samples}.

\subsection{Rendering and Optimization}
Rendering is performed using Blender, where the Stop Motion OBJ plugin \cite{githubGitHubNeverhood311StopmotionOBJ} is utilized to import and animate sequences of mesh files. This plugin enables seamless integration of mesh sequences into Blender, allowing for the efficient rendering of high-resolution synthetic videos. Our rendering pipeline supports two main perspectives: frontal and side views of the head. Each perspective contributes to the dataset's diversity, providing various viewpoints for improved model robustness.

To manage the large-scale rendering tasks, we employed a cluster system consisting of 20 nodes, each with 16 threads, totaling 320 concurrent threads. This distributed approach significantly enhances rendering efficiency, reducing the processing time for 8,600 videos to approximately 2 hours. As a side note, with more high-performance hardware, such as the Apple M1 chip, the render time per video can be reduced to approximately 1 minute,  and thus will significantly improve the scalability of the process.

This synthetic dataset generation pipeline offers a robust solution for augmenting data availability for pain recognition models while addressing ethical concerns related to real data collection. By combining advanced 3D modeling, texture mapping, and high-performance rendering techniques, our approach provides a scalable and privacy-preserving alternative to traditional dataset creation methods.

\section{Experimental Setup}\label{sec:model_design}
\begin{figure*}
    \centering
    \includegraphics[width=0.8\linewidth]{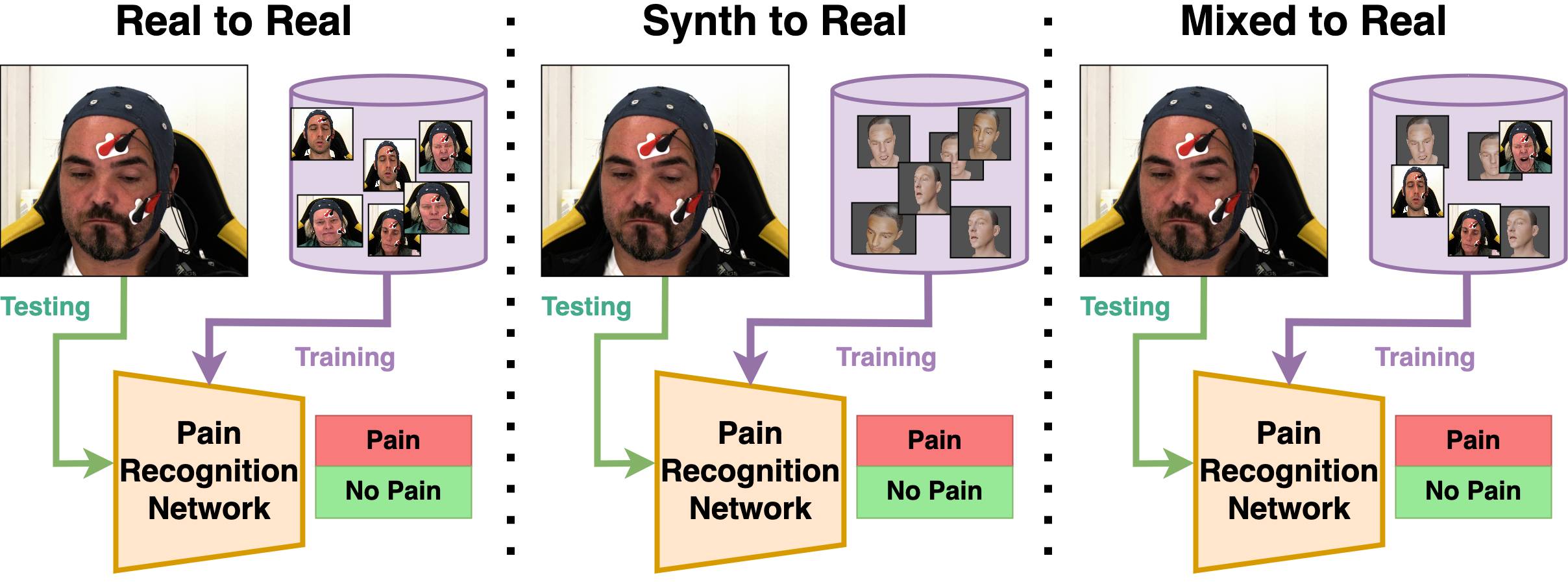}
    \caption{We consider three training configurations in our experiments: (1) using real data only (real to real), (2) using synthetic data exclusively (synth to real) and using both for training (mixed to real). }
    \label{fig:graphical_abstract}
\end{figure*}

To estimate the efficacy of our synthetic dataset generation pipeline, we consider a comparison of models trained on different combinations of real and synthetic data. This comparison assesses the potential of synthetic data to replicate real-world scenarios and enhance the performance and robustness of pain recognition models.
Let \( D_{R} \) represent the real data and \( D_{S} \) denote the synthetic data. Let \( D_{Tr} \),  \( D_{Val} \) and \( D_{Test} \) denote the training, validation and testing data that was used respectively.

All experiments consider the same setup of hyperparameters. The networks is a SlowFast-R50\cite{feichtenhofer2019slowfast} that is optimized for 100 epochs with a weighted BCE-loss using SGD with a batch size of 64, learning rate of 0.01, weight decay of 1e-5, and momentum of 0.9. During training, videos undergo random short side scale, random cropping, and random horizontal flip as augmentation. During validation, videos were subsampled, short-side scaled, center cropped. After every 10 epochs of training the SlowFast-R50 model, the model's progress is assessed using the \( D_{Val} \).  
To ensure representative training and validation datasets, data splitting followed a strategy proposed in previous research \cite{Biovidholdouteval}. This strategy aimed to maintain similar gender, age, and expressiveness distributions between the training and validation sets.

\subsection{Baseline Model (Real to Real)}

We use the BioVid Heat Pain Database \( D_{R} \) as the baseline dataset. The model trained solely on \( D_{R} \) provides a reference for further experiments with \( D_{Tr} \cup D_{Val} = D_{R}\),  \(  D_{Tr} \cap  D_{Val}=\emptyset\) with \(D_{Val}= D_{Test}\)
to stay consistent to the provided splits by \cite{Biovidholdouteval}.

\subsection{Synthetic Only Model (Synth to Real)}

In this setup, the model is trained exclusively on synthetic data \(  D_{Tr}, D_{Val} \in D_{S} \). Each test patient from the original dataset is augmented with multiple synthetic head textures. The final performance is then evaluated on the real data \( D_{Test} \in D_{R}\).  This configuration tests the feasibility of synthetic data in isolation and its potential to match or surpass the baseline performance.

\subsection{Synthetic and Real Model (Mixed to Real)}

To explore the benefits of combining synthetic and real data, we train a model on the combined dataset \( D_{C} = D_{R} \cup D_{S} \). This setup evaluates whether integrating synthetic data \( D_{S} \) with real data \( D_{R} \) enhances model performance beyond what is achievable with real data alone. 
\(  D_{Val} \in D_{S} \). Each test patient from the original dataset is augmented with multiple synthetic head textures. The final performance is then evaluated on the real data \( D_{Test} \in D_{R}\).
This evaluation aims to determine if the combination of synthetic and real data provides a substantial performance improvement.

\section{Results}\label{sec:results}

This section presents the results of our experiments, evaluating the performance of pain recognition models trained on different datasets: real, synthetic, and a combination of both. Subsequently, we provide an analysis of design decisions regarding the generation of synthetic data. The evaluation metrics used are Area Under the Receiver Operating Characteristic Curve (AUROC), F1-Score, and Accuracy. These metrics provide a comprehensive overview of the models' capabilities in handling the BioVid Heat Pain dataset, as detailed in Table \ref{tab:results}.

\subsection{Comparison of Training Configurations for Real-World Pain Recognition}

\begin{table}[b]
    \centering
    
    \caption{Pain recognition results on the \textbf{BioVid Heat Pain dataset}. We consider models trained on only real data (Real to Real), only synthetic data with 10 different textures per patient (Synth to Real), and a mix of both (Mixed to Real).}
    \begin{tabular}{|l||c|c|c|}
        \hline
        \textbf{Case} & \textbf{AUROC}  & \textbf{F1-Score} & \textbf{Accuracy} \\
        \hline\hline
         Real to Real & 0.741  & 0.666 & 0.654 \\
         Synth to Real& 0.581 & 0.8 & 0.666 \\
         Mixed to Real & 0.78 & 0.817 & 0.708\\
         \hline
    \end{tabular}
    \label{tab:results}
\end{table}

The baseline model, trained exclusively on the BioVid Heat Pain Database (Real to Real), achieved an AUROC of 0.741, an F1-Score of 0.666, and an Accuracy of 0.654. These results set a benchmark for evaluating pain recognition systems, illustrating the inherent challenges of using only real data, which is often limited in variability. The Synth to Real configuration, where the model was trained solely on synthetic data, demonstrated a lower AUROC of 0.581 but achieved a higher F1-Score of 0.800 and Accuracy of 0.666. This suggests that while synthetic data alone may struggle with overall class discrimination, it effectively captures the positive class, improving the balance between precision and recall. This indicates the potential of synthetic data to address class imbalance in pain recognition. Notably, the Mixed to Real model, which integrates both real and synthetic data, outperformed the individual datasets with an AUROC of 0.780, an F1-Score of 0.817, and an Accuracy of 0.708. The superior AUROC reflects an enhanced capacity to generalize across varied scenarios, leveraging the variability introduced by synthetic data. This configuration underscores the advantage of combining datasets, leading to improved generalization and robustness in pain recognition tasks. Overall, these findings affirm that incorporating synthetic data significantly enhances model performance, offering a practical solution to the limitations of traditional data collection methods and underscoring its potential as a valuable tool in advancing pain recognition technology.

\subsection{Aspect of Texture}

Table \ref{tab:Training result with synthetic data as validation set on the epoch with the highest AUROC-Score QUANTITY} evaluates model performance on synthetic datasets with varying texture complexity and patient representations. The \textit{10 Textures per Patient} configuration achieved the highest AUROC (0.655) and F1-Score (0.799), illustrating the positive impact of extensive texture diversity on pain recognition. 
The baseline \textit{Only Mesh} setup showed that models relying solely on mesh geometry achieved a lower AUROC of 0.624, highlighting the limitations of texture-free models. As texture diversity increased, there was a notable improvement in both AUROC and accuracy metrics, demonstrating that richer texture information  enhances the model's ability to capture nuanced pain indicators. Notably, the \textit{10 Textures per Patient} configuration resulted in the best overall performance, underscoring synthetic data capability to complement traditional pain recognition methods and improve generalization across varied scenarios.

\begin{table}[b]
    \centering
    
    \caption{Training result with \textbf{synthetic data as validation set} on the epoch with the highest AUROC-Score}
    \begin{tabular}{|l||c|c|c|c|c|}
        \hline
        \textbf{Research Case} & \textbf{AUROC} & \textbf{F1-Score} & \textbf{Accuracy}\\
        \hline\hline
         Only Mesh& 0.624 & 0.799 & 0.665\\
         1 Texture/patient& 0.643  & 0.776 & 0.667\\
         2 Textures/patient& 0.629 & 0.752 & 0.634\\
         3 Textures/patient& 0.637 & 0.745 & 0.647\\
         5 Textures/patient& 0.642 & 0.767 & 0.664\\
         10 Textures/patient& 0.655 & 0.799 & 0.668\\
         \hline
    \end{tabular}
    \label{tab:Training result with synthetic data as validation set on the epoch with the highest AUROC-Score QUANTITY}
\end{table}

\subsection{Aspect of View}

Table \ref{tab:Training result with synthetic data as validation set on the epoch with the highest AUROC-Score QUALITY} summarizes the performance of models trained on synthetic data across different view configurations. Both the Front View and Side View models yielded identical metrics: AUROC of 0.641, F1-Score of 0.786, and Accuracy of 0.658, indicating that each view independently offers a similar level of information for pain recognition, and thus are both helpful in their own way. The Multiple Views configuration using both views achieved the highest performance with an AUROC of 0.653, F1-Score of 0.799, and Accuracy of 0.666. This demonstrates the enhanced generalization capability afforded by combining diverse perspectives, suggesting that multi-view synthetic data significantly improves the model's ability to capture  pain-related expressions.

\begin{table}[H]
    \centering
    
    \caption{Training result with \textbf{synthetic data as validation set} on the epoch with the highest AUROC-Score}
    \begin{tabular}{|l||c|c|c|}
        \hline
        \textbf{Research Case} & \textbf{AUROC} & \textbf{F1-Score} & \textbf{Accuracy} \\
        \hline\hline
         Front View& 0.641 & 0.786 & 0.658\\
         Side View& 0.641 & 0.786 & 0.658\\
         Multiple Views& 0.653 & 0.799 & 0.666\\
         \hline
    \end{tabular}
    \label{tab:Training result with synthetic data as validation set on the epoch with the highest AUROC-Score QUALITY}
\end{table}

\section{Discussion and Conclusion}
Our work advances video-based pain recognition by introducing a pipeline for generating synthetic video data using deep learning techniques. Leveraging the BioVid Heat Pain Database, we combined the EMOCA repository for meshes, FFHQ-UV for textures, and Blender for rendering to create a diverse dataset with varied textures and angles. Our results show that while synthetic data alone has limitations, integrating it with real data significantly enhances pain recognition performance.

This approach addresses the ethical and privacy concerns associated with real-world pain data collection by offering a privacy-preserving alternative. Synthetic data generation not only circumvents privacy issues but also enables the inclusion of diverse demographics, enhancing the dataset's representativeness. 
However, models trained solely on synthetic data show limited performance improvements, highlighting the need for effective synthetic-to-real generalization. Future work should focus on enhancing this generalization, exploring advanced generative models, and incorporating additional data sources such as biomedical signals to improve the applicability of synthetic data in real-world scenarios. 
In conclusion, our study lays the groundwork for more efficient and ethical medical data collection, advancing pain recognition models while respecting patient privacy and promoting inclusivity of different appearances.

\paragraph{Acknowledgements.} A. Roitberg was supported by  DFG under Germany’s Excellence Strategy - EXC 2075 (SIMTECH).
This work received funding from ‘KITE’ (Plattform für KI-Translation Essen) from the REACT-EU initiative \\(\href{https://kite.ikim.nrw/}{https://kite.ikim.nrw/}, EFRE-0801977) and the Cancer Research Center Cologne Essen (CCCE).


\bibliographystyle{cas-model2-names}

\balance
\bibliography{cas-refs}



\end{document}